\pdfoutput=1

\documentclass[11pt]{article}

\usepackage[]{acl}

\usepackage{algorithm}
\usepackage{algpseudocode}
\usepackage{times}

\usepackage{float}
\usepackage{latexsym}
\usepackage{algorithm}
\usepackage{algpseudocode}

\usepackage[utf8]{inputenc}

\usepackage{microtype}

\usepackage{inconsolata}
\usepackage{CJKutf8}
\definecolor{ao(english)}{rgb}{0.0, 0.5, 0.0}
\usepackage{amsmath}
\usepackage{booktabs}
\usepackage{graphicx}

\usepackage{graphicx}
\usepackage{multirow}
\usepackage{adjustbox}
\usepackage{booktabs}

\usepackage{caption}
\usepackage{subcaption}

\definecolor{airforceblue}{rgb}{0.36, 0.54, 0.66}
\definecolor{blue(ncs)}{rgb}{0.0, 0.53, 0.74}
\definecolor{blush}{rgb}{0.87, 0.36, 0.51}
\definecolor{softgreen}{rgb}{0.7, 0.9, 0.7}
\definecolor{darkgreen}{rgb}{0.0, 0.5, 0.0} 
\definecolor{darkred}{rgb}{0.5, 0.0, 0.0} 

\usepackage{pifont}

\newcommand{\cmark}{\ding{51}}%
\newcommand{\xmark}{\ding{55}}%

\title{CrossIn: An Efficient Instruction Tuning Approach for Cross-Lingual Knowledge Alignment}

\author{
Geyu Lin\textsuperscript{$\heartsuit$}, 
Bin Wang\textsuperscript{$\heartsuit,\diamondsuit$}, 
Zhengyuan Liu\textsuperscript{$\heartsuit,\diamondsuit$}, 
Nancy F. Chen\textsuperscript{$\heartsuit,\diamondsuit,\dag$}
\\
\textsuperscript{$\heartsuit$}Institute for Infocomm Research (I$^2$R), A*STAR, Singapore\\
\textsuperscript{$\diamondsuit$}CNRS@CREATE, Singapore\\
\textsuperscript{$\dag$}Centre for Frontier AI Research (CFAR), A*STAR, Singapore\\
\texttt{lin\_geyu@i2r.a-star.edu.sg} \\
}

\begin{document}
\maketitle
\begin{abstract}

Multilingual proficiency in large language models (LLMs) presents a significant challenge due to the uneven distribution of training data and the English-centric focus during instruction tuning. To mitigate these issues, we introduce \texttt{CrossIn} (Cross-lingual Instruction Tuning), which utilizes two distinct types instruction tuning datasets: the Complex Task Dataset (CTD), rich in diverse, high-quality and logical tasks like math and coding, and the Linguistic Uniformity Dataset (LUD), consisting of easier-to-translate, linguistically uniform tasks. \texttt{CrossIn} merges cross-lingual instruction data from CTD with machine-translated data from LUD to reinforce knowledge alignment during instruction tuning. This strategy allows us to enhance reasoning capabilities without compromising knowledge consistency across languages. We also present a multi-task benchmark to evaluate \texttt{CrossIn}, with results showing substantial improvements in performance across languages and tasks. This demonstrates the benefits of integrating cross-lingual data and translation in enhancing multilingual consistency and accuracy during instruction tuning. \footnote{Code is available at https://github.com/Lingy12/CrossIn}
 
\end{abstract}

\section{Introduction}

    The advancement of large language models (LLMs) like ChatGPT \cite{achiam2023gpt} and Gemma \cite{team2023gemini} has been a game-changer in the field of natural language processing (NLP), revolutionizing tasks such as language generation and commonsense reasoning \cite{naveed2024comprehensive}. Nevertheless, most state-of-the-art LLMs are English-centric, and their performance on non-English languages is usually suboptimal, especially on languages that are dissimilar to English \cite{blevins2022language, mehrabi2022survey, gao2024multilingual}.
    This challenge mainly stems from the imbalanced distribution of multilingual data at both the pre-training and instruction tuning stages. The exposure bias toward major languages results in an imbalanced capability, where models excel in languages with plentiful data while under-performing in those with limited resources \cite{dac2023okapi, feng2023pretraining}. Bridging the language gap is a fundamental step to unlock the full potential of these general-purpose models and ensure that the benefits are accessible to people across the linguistic spectrum \cite{zhu2023multilingual}.

    Efforts to improve the multilingual capabilities of English-centric LLMs have involved continue pre-training using extensive language-specific datasets. Yet, mastering languages through additional pre-training could require vast amounts of data and significant computational resources \cite{le2022bloom}.
    On the other hand, despite the limited proportion of non-English data at the pre-training stage, their absolute volume builds a solid knowledge base of various languages. In each iteration, LLMs are exposed to samples in several languages simultaneously, and the compressed representation encourages models to share linguistic features and generalize across different languages \cite{le2022bloom}. However, this ability is not fully retained through the use of datasets that only include English in follow-up tuning steps.

    In this paper, we explore instruction tuning with two distinct datasets: the Complex Task Dataset (CTD), which contains a variety of high-quality, hard-to-translate tasks like math and coding, and the Linguistic Uniformity Dataset (LUD), characterized by easier-to-translate, linguistically uniform tasks. Tuning with CTD often leads to lower consistency due to model forgetting \cite{luo2024empirical}, while tuning with LUD may lead to lower reasoning capability due to its homogeneous nature.
    
    To address these issues, we propose a method that enhances both the consistency of cross-lingual instruction tuning and task performance accuracy. Our approach leverages advanced tuning strategies that exploit the logical structures within tasks, thereby improving logical reasoning and model effectiveness across different languages. This method not only balances the complexities associated with language-specific nuances but also enhances overall model performance in multilingual environments. Our results demonstrate substantial improvements in multilingual proficiency and task accuracy, advancing the capabilities of language models.

    To extensively evaluate the cross-lingual knowledge alignment \cite{qi2023crosslingual,wang2023seaeval}, we establish a benchmark of three tasks (i.e., reading comprehension, commonsense question-answering, and logic reasoning). Consistency is measured by analyzing an LLM's responses to the same question in different languages, and our benchmark encompasses multiple ability aspects and difficulty levels. Moreover, since exact match and F1 score cannot precisely evaluate system outputs in the generative setting, we unify all three tasks in a multiple-choice format for quantitative and reproducible evaluation.
    The experimental results demonstrate that our mixed cross-lingual tuning can significantly improve performance in all aspects (up to 40\% relative gain), followed by a detailed analysis of the influence of data quantity on language consistency and knowledge accuracy.

    The main contributions of our research are:
    
    \begin{itemize}
    \item \textbf{A Multi-faceted Benchmark.} We present a multi-lingual, multi-capability benchmark for assessing the cross-lingual knowledge consistency of language models. In particular, we build a parallel multiple-choice version of the XQuAD dataset~\cite{Artetxe:etal:2019} - Cross-XQuAD for machine comprehension, and combining it with commonsense QA and logic reasoning.
    
    \item \textbf{Mixed Cross-Lingual Instruction Tuning.} We introduce \texttt{CrossIn}, a cross-lingual instruction tuning approach aimed at aligning knowledge across languages to stimulate the model's full multilingual capability after pre-training. It offers a more reliable way of improving the model's capability without loss of consistency.
    
    \item \textbf{\texttt{CrossIn} Data Insights.} We conduct extensive experiments with representative LLMs on three tasks, and show the effectiveness of our proposed approach. We provide detailed analysis to study the optimal amount of cross-lingual data and the necessity of sample translation in enhancing models' cross-lingual consistency.
    \end{itemize}

\section{Related Work}

    \subsection{Multilingual Large Language Model}
        Multilingual Large Language Models (MLLMs) have experienced significant advancements in recent years. Recently, \citet{qin2024multilingual}, as a comprehensive review, summarizes various methodologies for training MLLMs. BLOOM~\cite{le2022bloom}, Jais~\cite{sengupta2023jais}, and Sailor~\cite{dou2024sailor} are representative models that target improved multilingualism in the pretraining stage. For fine-tuning, ChatGLM employs a reward model trained under a multilingual setting~\cite{zeng2022glm}, while the x-LLM utilizes a translated version of the  Alpaca dataset, combined with supervised translation data and instruction finetuning, to enhance the model's multilingual capabilities \cite{zhu2023extrapolating}.

      Instruction tuning using English datasets has demonstrated the potential to extend zero-shot capabilities across multiple languages \cite{wei2022finetuned, chung2022scaling}, however, they do have some limitation which is discussed in \ref{sec:data_5_2}. Previous research supports the notion that utilizing training sets composed of diverse languages can significantly enhance cross-lingual generalization \cite{muennighoff2023crosslingual, kew2023turning, shaham2024multilingual}. Building on these insights, our work focuses on enhancing multilingual consistency by specifically targeting instruction finetuning. By optimizing the instruction processing mechanism, we aim to ensure better alignment across different languages during instruction tuning phase. 
            
    \begin{figure*}[t]
    \centering
    \begin{subfigure}{0.38\textwidth}
      \centering
      \includegraphics[width=\linewidth]{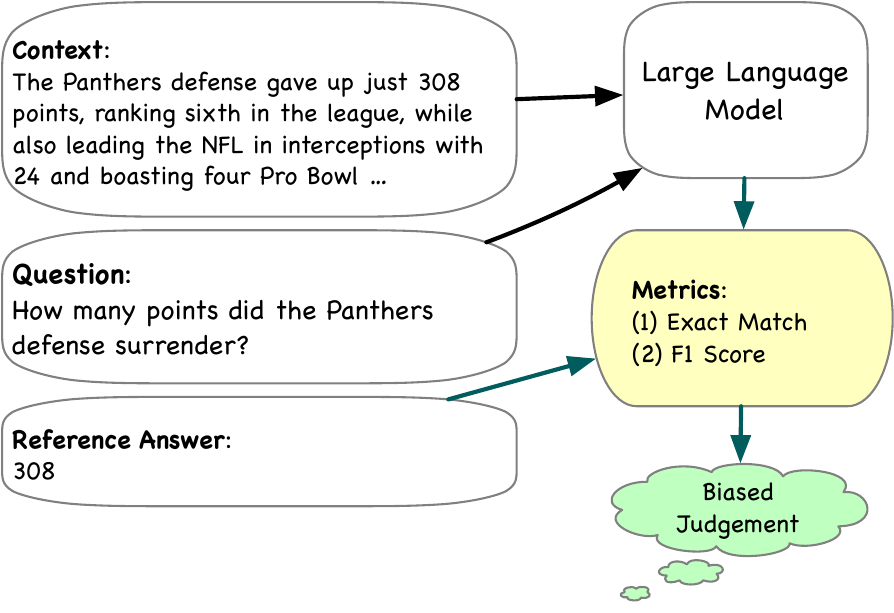}
      \caption{Original XQuAD Dataset}
      \label{fig:cross-xquad-sub1}
    \end{subfigure}%
    \hfill
    \begin{subfigure}{0.56\textwidth}
      \centering
      \includegraphics[width=\linewidth]{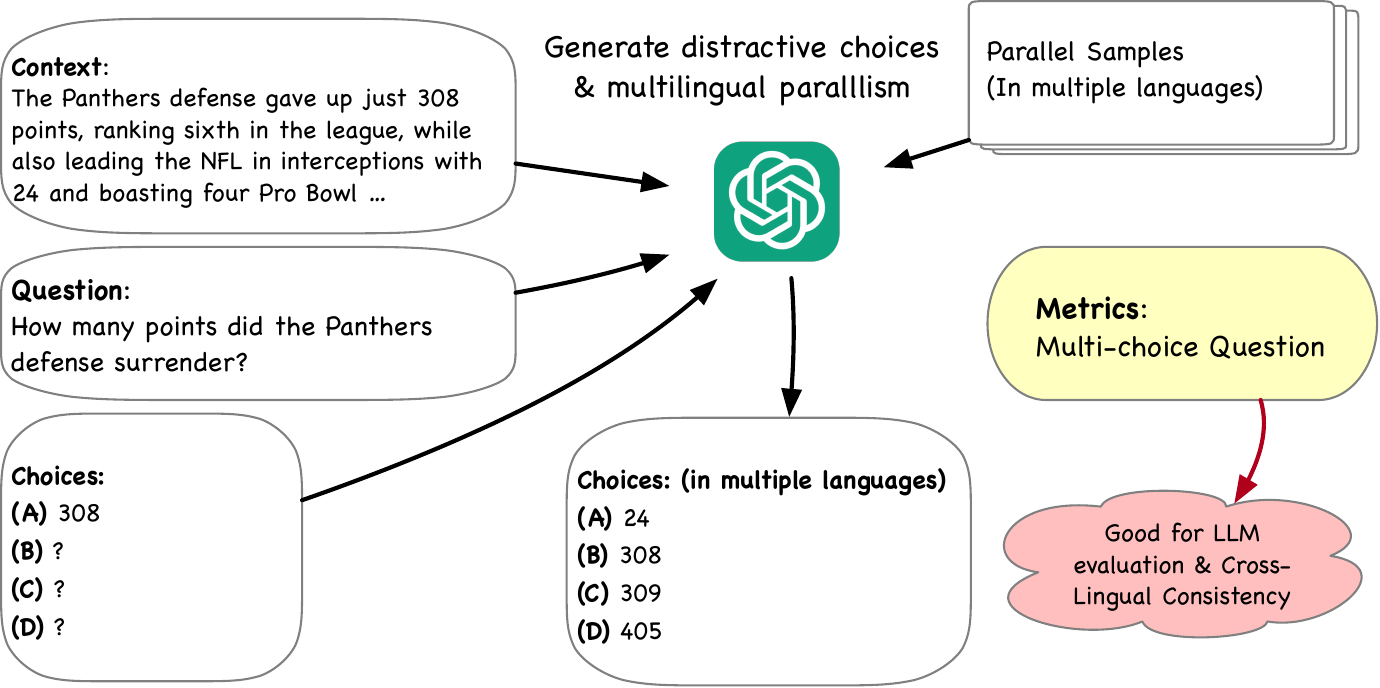}
      \caption{Cross-XQuAD Dataset Creation}
      \label{fig:cross-xquad-sub2}
    \end{subfigure}
    \caption{An illustration of the dataset construction process of the Cross-XQuAD dataset. The original XQuAD dataset, although multilingual, is not adapted specifically to evaluate LLMs and their cross-lingual consistency.}
    \label{fig:cross-xquad}
    \end{figure*}

    \subsection{Multilingual Evaluation Benchmark}

        Evaluating the multilingual capabilities of LLMs is crucial for their global applicability, as it ensures that these models can understand and generate text effectively across different languages. Benchmarks such as MMLU~\cite{hendrycks2021measuring}, TruthfulQA~\cite{lin2021truthfulqa} have been developed to access the general capability of the LLMs in English. XQuAD~\cite{Artetxe:etal:2019} and MLQA~\cite{lewis2019mlqa} are popular extractive question-answering datasets that have been developed to evaluate the models' multilingual performance. However, they focus on language-specific performance without considering the knowledge-sharing capabilities.
        Recently, Cross-MMLU and Cross-LogiQA~\cite{wang2023seaeval} are proposed to assess the multilingual capability of LLMs with an emphasis on cross-lingual consistency. However, the number of samples is limited which could generally lead to less stable evaluation results.
    
\begin{table*}[t]
    \centering
    \begin{adjustbox}{width=1.0\textwidth,center}
    \begin{tabular}{ l | c | c | c | c }
    \toprule
        \textbf{Dataset} & \textbf{MCQs} & \textbf{Number of Samples} & \textbf{Supported Language} & \textbf{Consistency Metric}\\\hline
        \multirow{1}{*}{MLQA ~\cite{lewis2019mlqa}} & \multirow{1}{*}{\xmark} & \multirow{1}{*}{5,500 (36$\times$)} & 7 - Eng, Zho, Spa, Vie, ... & NA \\ 
        \multirow{1}{*}{XQuAD~\cite{Artetxe:etal:2019}} & \multirow{1}{*}{\xmark} & \multirow{1}{*}{1,190 (7.9$\times$)} & 10 - Eng, Zho, Spa, Vie, ... & \multirow{1}{*}{NA} \\
        \hline \hline
        Cross-MMLU~\cite{wang2023seaeval} & \cmark & 150 (1$\times$) & 7 - Eng, Zho, Spa, Vie, ... & \cmark \\
        Cross-LogiQA \cite{wang2023seaeval} & \cmark & 176 (1.2$\times$) &  7 - Eng, Zho, Spa, Vie, ... & \cmark \\ \hline
        Cross-XQuAD (ours) & \cmark & 1,190 (7.9$\times$) & 4 - Eng, Zho, Spa, Vie & \cmark \\ 
    \bottomrule
    \end{tabular}
    \end{adjustbox}
    \caption{A list of multilingual datasets. Multi-choice questions (MCQs) are more suitable for quantitative evaluation of large language models and evaluation for multilingual consistency. Traditional metrics such as the F1 score or Exact Match for extractive question answering can introduce unintended biases in evaluating large language models.}
    \label{tab:multilingual-benchmark}
\end{table*}

\section{Cross-Lingual Consistency Benchmark}
\label{sec:benchmark}

    Since traditional multilingual evaluations often fail to cater specifically to LLMs or overlook the assessment of cross-lingual consistency in multilingual contexts, in this section, we present a targeted multilingual evaluation benchmark for cross-lingual knowledge alignment.

    \subsection{Datasets and Metrics}

        Even though there are multilingual evaluation datasets with parallel samples including MLQA~\cite{lewis2019mlqa} and XQuAD~\cite{Artetxe:etal:2019}, they are tailored for supervised extractive question-answering tasks and are unsuitable for less structured outputs of LLMs~\cite{schuster2023semqa}. Therefore, recently, two evaluation datasets have been developed for multilingual evaluation with cross-lingual consistency measures~\cite{wang2023seaeval}. Specifically, Cross-MMLU and Cross-LogiQA are designed to use multiple-choice questions, presenting parallel samples to assess the knowledge alignment capability of LLMs. These datasets focus on commonsense question answering and logical reasoning. However, as they are crafted by humans, the number of parallel samples they offer is relatively limited due to the high cost of human labor involved. This limitation could lead to less robust evaluation results.

        Considering this, in our work, we enhance the cross-lingual consistency evaluation benchmark by introducing another task type: reading comprehension. Furthermore, we utilize existing high-quality parallel datasets to automatically generate new ones that are tailored for LLM evaluation. Table~\ref{tab:multilingual-benchmark} summarizes the complete benchmark.

        For evaluation metrics, we leverage the same concept as presented in~\citet{wang2023seaeval}. In addition to assessing the overall accuracy of each language, we also integrate cross-lingual consistency metrics, measured by ``Consistency'' and ``AC3''. The consistency score is designed to determine whether the model provides consistent responses to parallel questions across different languages. A higher consistency score suggests that LLMs can apply common knowledge across languages and deliver uniform responses, regardless of correctness. Specifically, for the Cross-XQuAD dataset that spans four languages, the multilingual consistency metric is defined as         
        \begin{equation}
        M_{\{l_1, l_2, ..., l_s\}} = \frac{\sum_{i=1}^{N} 1\{a_{i}^{l_1} = a_{i}^{l_2} = ... = a_{i}^{l_s} \}}{N}
        \end{equation}
        where $a_i^{l_s}$ is the answer for sample index $i$ from language $s$. Then, the consistency is computed as: 
        
        \begin{equation}
        \textit{Consistency}_s = \frac{\sum_{\{l_1, l_2, ..., l_s\} \in C(s, g_i)} M_{\{l_1, l_2, ..., l_s\}}}{C^{s}_4}
        \end{equation}

        Similar to~\citet{wang2023seaeval}, we use $s=3$ as the default tolerant for consistency metrics, where the consistency between any three languages is computed.
        AC3 enhances the traditional accuracy metric by incorporating consistency, offering a more comprehensive evaluation. This approach is adopted because relying solely on consistency or accuracy does not yield a robust assessment.
        \begin{equation}
        AC3_s = 2 \cdot \frac{\textit{Accuracy} \cdot \textit{Consistency}_s}{\textit{Accuracy} + \textit{Consistency}_s}
        \end{equation}
        
        By converting the datasets into MCQ (Multiple Choice Question) format, we can better quantify the model's ability to select the correct answer from a set of options, thereby offering a clearer measure of its understanding and reasoning capabilities.          

    \subsection{Cross-XQuAD Construction}

        Figure~\ref{fig:cross-xquad} indicates the process of constructing the Cross-XQuAD dataset from the original XQuAD dataset. It involves three steps, 1) English MCQ construction with distractive choices, 2) Parallel MCQ construction, and 3) Post-processing and quality check.

        First, the original ground-truth answer from the XQuAD dataset can directly be used as the correction choice. As the XQuAD is for an extractive question-answer task, we extract the incorrect options from the provided context corpus as much as possible. Otherwise, the solution would be highly trivial with simple matching techniques. To achieve this, we prompt \emph{ChatGPT-3.5} to get the other three choices as shown in Figure~\ref{fig:cross-xquad-sub2}.

        Second, using the prepared English sample as a base, we prompt the generation of equivalent samples in the other languages. We discovered that direct translation without specific context can result in deviated interpretations due to polysemy, potentially leading to a biased evaluation. To counter this, we prompt the model with the English sample alongside its contextual counterpart in the target language to generate new samples. This approach has resulted in samples that are highly aligned across multiple languages.

        Third, although LLMs can perform as a reasonable automated method for creating parallel samples \cite{li2023coannotating}, we found that human intervention is essential to ensure higher accuracy. Consequently, each constructed sample undergoes a round of human review to confirm its integrity.
        
        Following the above procedure, we construct the Cross-XQuAD dataset with 1,190 parallel samples in four languages which results in 4,760 samples in total. It is by far the largest multilingual evaluation dataset with cross-lingual consistency assessment capabilities.

\begin{table*}[ht]
    \centering
    \small
    \begin{adjustbox}{width=1.0\textwidth,center}
    \begin{tabular}{ l | p{7.5cm} | p{7.5cm}  }
    \toprule
     \textbf{Type} & \textbf{Instruction} & \textbf{Output} 
     \\ \midrule

     Original Sample
    & \emph{Explain what a "food desert" is and why they are a problem.}
    & \emph{A "food desert" is an area with little or no access to affordable, healthy food options....}
     \\ \midrule
     \texttt{CrossIn$_{en2x}$}
     & \emph{Explain what a "food desert" is and why they are a problem.} 
     & \begin{CJK*}{UTF8}{gbsn} “食物沙漠” 是一个几乎没有或根本没有负担得起的健康食品选择的地区... \end{CJK*} 
     \\ 
     
     \midrule
     
     \texttt{CrossIn$_{x2en}$}
     & \begin{CJK*}{UTF8}{gbsn} 解释什么是 “食物沙漠”，以及为什么它们是一个问题\end{CJK*} 
     & \emph{A "food desert" is an area with little or no access to affordable, healthy food options....}
     \\ 
     
     \midrule
     
     \texttt{CrossIn$_{x2x}$} 
     & \begin{CJK*}{UTF8}{gbsn} 解释什么是 “食物沙漠”，以及为什么它们是一个问题\end{CJK*} 
     & \emph{Un "desierto alimentario" es un área con poco o ningún}
     \\ 
     \texttt{(zho-spa)} & & \emph{acceso a opciones de alimentos saludables y asequibles...}
     \\
    \midrule
    
     \texttt{CrossIn$_{x2x}$} 
     & Explain what a "food desert" is and why they are a problem.
     & \emph{Un "desierto alimentario" es un área con poco o ningún}
     \\ 
     \texttt{(eng-spa)} & & \emph{acceso a opciones de alimentos saludables y asequibles...}
     \\
     \midrule

     \texttt{CrossIn$_{x2x}$} 
     & Explique qué es un "desierto alimentario" y por qué son un 
     & \begin{CJK*}{UTF8}{gbsn} “食物沙漠” 是一个几乎没有或根本没有负担得起的健\end{CJK*} 
     \\ 
     \texttt{(spa-zho)} & problema. &  \begin{CJK*}{UTF8}{gbsn} 康食品选择的地区... \end{CJK*} 
     \\
     \midrule
    
     Translation
     & \emph{Translate the following sentence into English.} 
     
     \begin{CJK*}{UTF8}{gbsn}  解释什么是 “食物沙漠”，以及为什么它们是一个问题\end{CJK*}
     & \emph{Explain what a "food desert" is and why they are a problem.}
     \\
     \bottomrule
    \end{tabular}
    \end{adjustbox}
    \caption{One example from the Alpaca dataset. It is further transformed into cross-lingual instruction tuning datasets and translation tasks. 
    }
    \label{tab:cross-lingual-alignment-example}
\end{table*}

\section{\texttt{CrossIn} Method}

    To address language imbalances in English-centric LLM pre-training and fine-tuning, we explore cross-lingual instruction tuning. Traditional methods, primarily using monolingual (English) samples, limit broad multilingual engagement \cite{zhu2023extrapolating}. Our approach, \texttt{CrossIn}, integrates mixed language compositions at the sample level to enhance both task-solving abilities and multilingual proficiency by utilizing the shared compressed representation space across languages. This strategy effectively combines the simplicity of linguistic datasets like Alpaca with the complex, hard-to-translate tasks in Platypus, enhancing language-level generalization and boosting the model's task solving capabilities.

    The training data can be divided into three main aspects: \textbf{\texttt{Base}}, \textbf{\texttt{CrossIn}}, \textbf{\texttt{Trans}}.
    
    \begin{itemize}
        \item \textbf{\texttt{Base}}: This section covers the foundational instruction tuning datasets that the model uses to acquire all basic capabilities. English datasets, which are the most resource-rich and of the highest quality, can be classified as the Complex Task Dataset (CTD).

        \item \textbf{\texttt{CrossIn}}: This component consists of cross-lingual instruction tuning datasets, where instructions and outputs are presented in two different languages. This segment should be sourced from a dataset that exclusively contains pure linguistic content, making it easy to translate, and can be classified as the Linguistic Uniformity Dataset (LUD).
        
        \item \textbf{\texttt{Trans}}: It consists of translation pairs for instructions. We hypothesize that if the model concurrently learns these translation tasks, it could facilitate the transfer of knowledge between languages.

    \end{itemize}

    For \textbf{\texttt{Base}}, we leverage existing datasets that has diverse task including math and code, where we use Open-Platypus as source \cite{lee2023platypus}. we create the \textbf{\texttt{CrossIn}} and \textbf{\texttt{Trans}} datasets, where we use the Alpaca \cite{alpaca} dataset as the source. Examples are shown in Table~\ref{tab:cross-lingual-alignment-example}.

    For \textbf{\texttt{CrossIn}} dataset, we create three variants as the following recipes:

    \begin{itemize}
    \item \texttt{CrossIn$_{en2x}$}: Instructions are provided in English, and we choose the output language randomly. Given the rich prior knowledge available in English, this approach aims to transfer English knowledge to other languages.
    
    \item \texttt{CrossIn$_{x2en}$}: Instruction language is chosen randomly, and output is fixed in English. This approach aims to unify multilingual instructions into responses centered around English.
    
    \item \texttt{CrossIn$_{x2x}$}: The languages for both the instruction and the output are selected randomly. This approach seeks to facilitate bi-directional alignment across all languages.
    \end{itemize}

    \begin{algorithm}[t]
        \caption{\texttt{CrossIn$_{x2x}$} with translation}
        \begin{algorithmic}
        \State $\mathcal{S} \gets \text{Total number of samples}$
        \State $\mathcal{L} \gets \{\text{"English", "Spanish", "Chinese",}$
        \State \quad\quad\quad$\text{"Vietnamese"}\}$
        \State $\mathcal{D} \gets \text{Seed Parallel Instructions Dataset}$
        \State $\mathcal{C} \gets \emptyset$
        \State $\mathcal{T} \gets \emptyset$
        \State $t_p \gets \text{Translation Prompt}$
        \For{$i \gets 1$ to $\mathcal{S}$}
            \State $s \gets \text{Random sample from } \mathcal{D}$
            \State $l_{in},l_{ot} \gets \text{Random sample from } \mathcal{L}$ 
            \State $\mathcal{C} \gets \mathcal{C} \cup (\mathcal{D}[l_{in}][s], \mathcal{D}[l_{ot}][s])$
            \State $l_t \gets \text{Random sample from } \mathcal{L}$ 
            \State $\mathcal{T} \gets \mathcal{T} \cup (t_p, D[l_t][s], D[``English"][s])$
        \EndFor
        \end{algorithmic}
        \label{algorithm:1}
    \end{algorithm}

    Previous work shows that incorporating sample translation helps map English to other languages, allowing the model to generalize English knowledge in a broader space \cite{zhu2023extrapolating}. For an extensive comparison, we also investigate how adding a separate translation task might enhance the multilingual abilities of LLMs, compared with using cross-lingual instruction tuning alone. More specifically, aside from the \texttt{CrossIn} data, we add a direct translation task of instructions from English to other languages. The influence on model performance of additional instruction translation is discussed in Section \ref{sec:ablation_study}.

    Algorithm~\ref{algorithm:1} illustrates the complete algorithm to create \texttt{CrossIn$_{x2x}$} with translation dataset, where $\mathcal{S}$ is the desired number of samples to be added with the \textbf{\texttt{Base}}. $\mathcal{C}$, $\mathcal{T}$, $l_{in}$ indicate \textbf{\texttt{CrossIn}}, \textbf{\texttt{Trans}} and the sampled language, respectively.  


    \begin{table*}[t]
        \centering
        \begin{adjustbox}{width=1.00\textwidth,center}
        \begin{tabular}{l | c  c  c | c  c  c | c  c  c }
        \toprule
         \multirow{2}{*}{\textbf{Models}} & \multicolumn{3}{c|}{\textbf{Cross-XQuAD}} & \multicolumn{3}{c|}{\textbf{Cross-MMLU}} & \multicolumn{3}{c}{\textbf{Cross-LogiQA}} \\  \cline{2-10}
         & Acc & Consis & AC3  & Acc & Consis & AC3  &  Acc & Consis & AC3  \\\hline\hline
         \multicolumn{1}{l}{{\textbf{General LLMs}}} \\\hline\hline
         
         \emph{ChatGPT-3.5} & 90.6 & 83.7 & 87.0 & 66.8 & 51.8 & 58.4 & 53.3 & 40.5 & 46.0 \\
         \emph{LLaMA-2-7B-Chat}~\cite{touvron2023llama} & 74.9 & 67.5 & 71.1 & 40.1 & 42.0 & 41.1 & 36.8 & 43.5 & 39.9 \\
         \emph{Mistral-7B-Instruct-v0.2}~\cite{jiang2023mistral} & 84.6 & 72.2 & 77.9 & 49.0 & 26.2 & 34.1 & 46.0 & 38.5 & 41.9 \\
         \emph{LLaMA-7B}~\cite{touvron2023llama-1} & 40.3 & 21.5 & 28.0 & 29.8 & 27.8 & 28.8 & 27.6 & 23.0 & 25.1 \\
         \emph{m-LLaMA-7B}~\cite{zhu2023extrapolating} & 46.8 & 41.1 & 43.8 & 26.7 & 22.3 & 24.3 & 28.1 & 22.0 & 24.7 \\
         \hline \hline

        \multicolumn{2}{l}{\textbf{Base Model}: \emph{Gemma-2B}~\cite{team2024gemma}} \\\hline\hline
         
        \emph{Tuning w/ Alpaca} & 42.0 & 49.7 & 45.5 & 36.0 & \textbf{59.8} & 45.0 & 28.3 & \textbf{63.8} & 39.2 \\
        \emph{Tuning w/ Platypus} & \textbf{60.8} & 55.8 & 58.2 & 36.5 & 29.7 & 32.7 & 36.4 & 47.9 & 41.3 \\\hline
         \texttt{CrossIn$_{en2x}$}  & 60.1 & 62.8 & \textbf{61.5} & 39.2 & 43.0 & 41.0 & 39.5 & 37.8 & 38.6 \\
         \texttt{CrossIn$_{x2en}$}  & 54.2 & \textbf{64.7} & 59.0 & \textbf{41.2} & 57.8 & \textbf{48.1} & 36.8 & 48.3 & 41.8 \\
         \texttt{CrossIn$_{x2x}$}  & 53.3 & 64.3 & 58.3 & 37.0 & 54.5 & 44.1 & \textbf{39.6} & 46.2 & \textbf{42.6}\\
         \hline\hline

         \hline\hline

        \multicolumn{2}{l}{\textbf{Base Model}: \emph{Mistral-7B-v0.1}~\cite{jiang2023mistral}} \\\hline\hline
          
    
         \emph{Tuning w/ Alpaca} & 62.2 & 52.9 & 57.2 & 36.2 & 43.5 & 39.5 & 35.7 & 33.8 & 34.7 \\
         \emph{Tuning w/ Platypus} & 61.1 & 33.2 & 43.0 & 38.8 & 20.2 & 26.5 & 47.9 & 29.8 & 36.8 \\ \hline
         \texttt{CrossIn$_{en2x}$} & 74,9 & 64.0 & 69.0 & 41.0 & 41.5 & 41.2 & 44.6 & 40.1 & 42.2 \\
         \texttt{CrossIn$_{x2en}$}  & 77.4 & 63.8 & 69.9 & 34.8 & \textbf{47.2} & 40.1 & 45.3 & 42.5 & 43.8\\
         \texttt{CrossIn$_{x2x}$} & \textbf{78.6} & \textbf{67.9} & \textbf{72.9} & \textbf{41.0} & 42.3 & \textbf{41.7} & \textbf{48.9} & \textbf{48.3} & \textbf{48.6} \\

          \hline\hline
      \multicolumn{2}{l}{\textbf{Base Model}: \emph{LLaMA-3-8B}~\cite{team2024llama3}} \\\hline\hline
         
        \emph{Tuning w/ Alpaca} & 85.7 & 79.9 & 82.7 & 51.2 & \textbf{42.3} & 46.3 & 42.3 & 40.2 & 41.2 \\
        \emph{Tuning w/ Platypus} & 86.8 & 75.6 & 80.8 & 53.0 & 32.8 & 40.5 & 56.5 & 48.2 & 52.0 \\\hline
         \texttt{CrossIn$_{en2x}$}  & 87.3 & 76.7 & 81.7 & \textbf{53.8} & 41.0 & \textbf{46.5} & 57.9 & 48.4 & 52.8 \\
         \texttt{CrossIn$_{x2en}$}  & 88.7 & 78.7 & 83.4 & 53.8 & 32.0 & 40.1 & 60.3 & 49.3 & 54.3 \\
         \texttt{CrossIn$_{x2x}$}  & \textbf{88.7} & \textbf{80.7} & \textbf{84.5} & 52.0 & 33.5 & 40.7 & \textbf{59.5} & \textbf{51.4} & \textbf{55.2} \\
         
        \bottomrule
        \end{tabular}
        \end{adjustbox}
        \caption{Experimental results on three cross-lingual consistency datasets: Cross-XQuAD, Cross-MMLU, Cross-LogiQA. Three metrics presented are Accuracy (ACC), Consistency (Consis), and AC3 as introduced in Section~\ref{sec:benchmark}.} 
        \label{tab:results}
    \end{table*}

\section{Experiments}

    \subsection{Experimental Setting}

        In our experiments, we selected four languages: English, Chinese, Vietnamese, and Spanish across all three datasets. We utilized two representative open LLM as base model: \emph{Mistral-7B-v0.1}~\cite{jiang2023mistral}, \emph{Gemma-2B}~\cite{team2024gemma} and \emph{LLaMA-3-8B}~\cite{team2024llama3}. For base models, we employed the Platypus~\cite{lee2023platypus} corpus as the \textbf{\texttt{Base}} dataset for instruction tuning, since previous work shows that it can enable models' higher diverse and robust generalization capabilities than the Alpaca dataset.

        For the \textbf{\texttt{CrossIn}} instruction tuning data, we utilize the Alpaca~\cite{alpaca} corpus as the seed dataset. This dataset is expanded into a multilingual format to four languages using an off-the-shelf translation engine, producing a total of (52k$\times$4) samples. From the enriched datasets, both the \textbf{\texttt{CrossIn}} and \textbf{\texttt{Trans}} parts can be formulated in a variant number of samples. While the Alpaca dataset lacks the complex problem-solving capabilities of the \textbf{\texttt{Base}} set from Platypus, it contains English instructions without complex elements like coding and math, which results in a higher translation quality. Meantime, this setup allows us to investigate whether a dataset of simple instructions can adequately support effective knowledge alignment across languages.

        In model training, we leverage LoRA~\cite{hu2022lora} with $rank=64$ as a parameter-efficient way to train LLMs. For fair comparison, we fine-tune base models with either the Platypus or Alpaca dataset with the same set of hyperparameters. Besides, following standard benchmarks, we also compared several representative general-purpose LLMs including \emph{ChatGPT-3.5}, \emph{LLaMA-2-7B-Chat}, \emph{Mistral-7B-Instruct-v0.2}. Models from previous work \cite{zhu2023extrapolating} 
        \emph{m-LLaMA-7B} and its base model, \emph{LLaMA-7B}.

    \subsection{Main Results and Analysis}
    \label{sec:data_5_2}

        Table~\ref{tab:results} shows the benchmark results of current general LLMs and models tuned with Alpaca, Platypus and different \texttt{CrossIn} variants. Our findings can be summarized as follows.

        \noindent\textbf{English-centric instruction tuning is limited.} We analyzed the performance of base models fine-tuned on LUD (Alpaca) and CTD (Platypus) respectively. Our findings indicate that models exhibit distinct characteristics depending on the instruction tuning corpus, especially on logical reasoning benchmarks. Fine-tuning with Platypus results in higher accuracy, potentially due to the diversity of tasks in the dataset. Conversely, models fine-tuned with Alpaca shows a higher consistency across most benchmark datasets, albeit with marginally lower accuracy especially for Cross-LogiQA which is focusing on logical reasoning. These observations suggest that Alpaca may be less effective than Platypus in augmenting LLMs with task-solving and reasoning. In addition, fusing a wide range of knowledge in English could potentially lead to a forgetting of information in other languages, thus affect the consistency. This results show a trade-off between accuracy and consistency from fine-tuning on different English-centric instruction tuning corpora. We aim to bridge the gap of both datasets with our method, thereby enhancing both accuracy and consistency.

        \noindent\textbf{Monolingual Mixture is not effective enough} \emph{m-LLaMA-7B} which uses mixture of multiple monolingual data with translation demonstrated some improvements over \emph{LLaMA-7B} in the Cross-XQuAD dataset, but it only managed to achieve similar results on the Cross-MMLU and Cross-LogiQA. This suggests that a purely monolingual data mix may not be adequate for training models on complex multilingual tasks, highlighting the importance of our proposed approach.

        \begin{figure}[t!]
        \centering
        \includegraphics[width=0.53\textwidth]{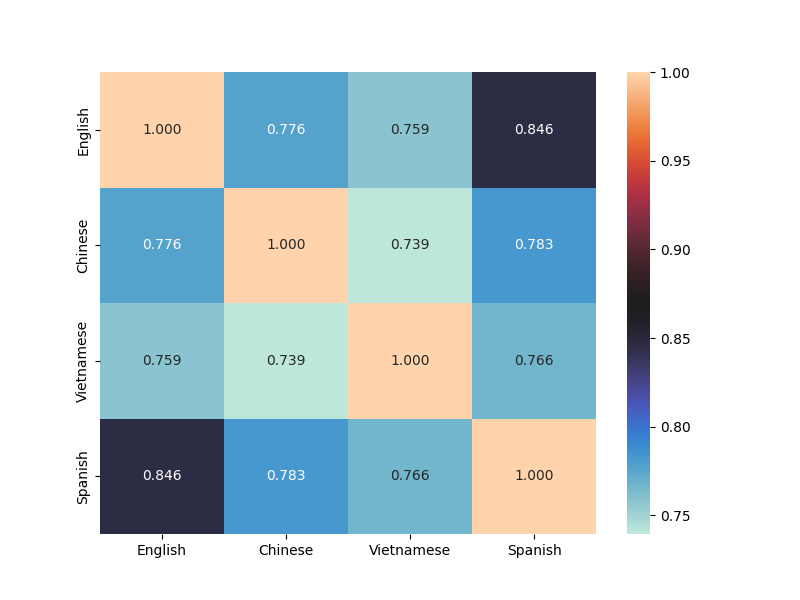}
        \caption{Consistency score between languages on Cross-XQuAD with \texttt{CrossIn$_{x2x}$} method}
        \label{fig:lang_heatmap}
        \vspace{-0.2cm}
        \end{figure}
        
        \noindent\textbf{\texttt{CrossIn} is simple but effective} We further review the results from our \texttt{CrossIn} instruction tuning method, which leverages the strengths of both the English-centric Platypus and the Multilingual Alpaca datasets. By implementing the \texttt{CrossIn} augmentation, we successfully raised the AC3 score by 30\% on the Cross-XQuAD benchmark and by about 12\% on both the Cross-MMLU and Cross-LogiQA testsets. This improvement was achieved using the \texttt{CrossIn$_{x2x}$} approach with the Mistral-7B-v0.1 as the foundational model. Enhancements were evident in the model's accuracy and consistency across various languages, contributing to the higher AC3 scores. Our results demonstrate the efficacy of the \texttt{CrossIn} method in enhancing the model's knowledge consistency and logical capabilities. \texttt{CrossIn} achieves the highest scores in accuracy, consistency, and AC3 in the Cross-LogiQA dataset, underscoring its robust multilingual logical reasoning capabilities. 
        
        \noindent\textbf{Language discrepancy affects consistency.}
        We explore the consistency scores across language pairs as depicted in Figure \ref{fig:lang_heatmap}. Spanish and English show the highest consistency, likely due to linguistic similarities. In contrast, Chinese and Vietnamese have the lowest consistency, possibly because of their distinct character sets and language bias during pre-training. Specifically, Vietnamese, often considered a low-resource language in pre-training, exhibits the least consistency with English. This highlights the need to diversify training data for language models to ensure equitable and effective representation of typically underrepresented languages.

    \begin{figure}[t!]
        \centering
        \includegraphics[width=1\linewidth]{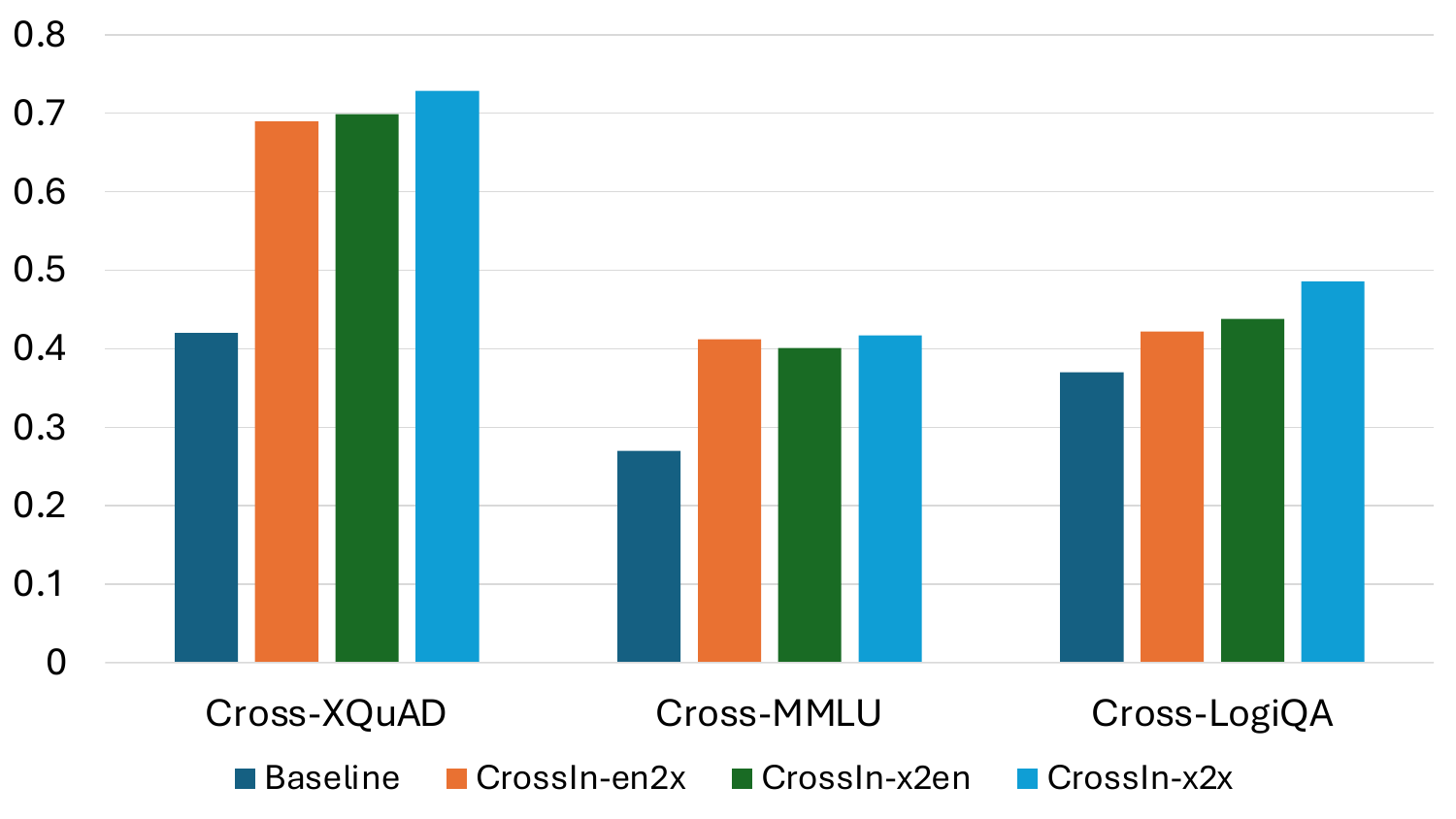}
        \caption{Results of different cross-lingual instruction tuning methods compared with baseline.}
        \label{fig:alignment}
    \end{figure}

    \subsection{Ablation Study}
    \label{sec:ablation_study}

         We conduct three comprehensive ablation studies to systematically assess the effects of various data formations, the integration of translation data, and the influence of different alignment dataset sizes on the performance of our models, aiming to identify key factors that enhance or inhibit their effectiveness.

        \noindent\textbf{Data Formulation Comparison.} Figure~\ref{fig:alignment} shows the AC3 scores from three tests when the language backbone is the Mistral-7B-v0.1. The results make it clear that methods designed for cross-lingual instructions work better than the basic method, which only uses English-centric instruction tuning data from Platypus or Alpaca. In particular, the \texttt{CrossIn}$_{x2x}$ method does much better than the \texttt{CrossIn}$_{en2x}$ and \texttt{CrossIn}$_{x2en}$ methods. This suggests that fully mixing multiple languages (\texttt{CrossIn}$_{x2x}$) can make the most of what the Mistral-7B-v0.1 model offers by effectively using data from different languages. The mixed composition in training examples seems to help the model understand and apply knowledge from one language to another, leading to more accurate and consistent results.

        \begin{figure}[t]
        \centering
        \includegraphics[width=1\linewidth]{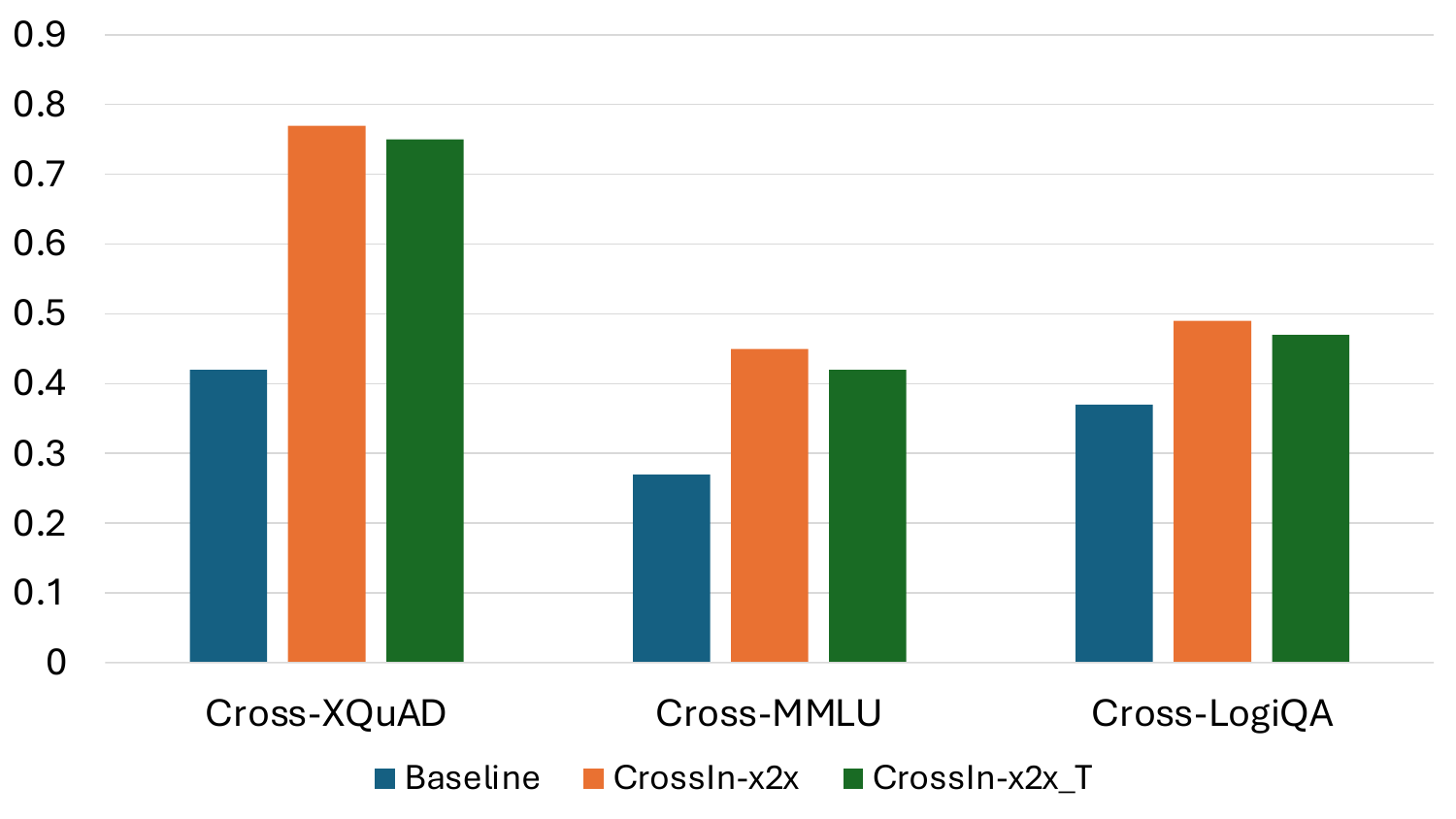}
        \caption{Comparison of AC3 score of adding translation data in cross-lingual instruction tuning.}
        \label{fig:translation}
        \end{figure}

        \noindent\textbf{Efficacy of Translation Data.} 
        Figure \ref{fig:translation} compares the performance of the \texttt{CrossIn}$_{x2x}$ method with the \texttt{CrossIn}$_{x2x\_T}$ strategy, which adds translations to the Alpaca samples (as described in Algorithm \ref{algorithm:1}). The experimental results indicate that additional translation pairs does not bring performance gains. We speculate that this is because tasks included in our benchmark focus on understanding and reasoning, and the cross-lingual instruction tuning approach stimulate both of them under a multilingual setting. Additionally, the translations used here may be too basic, especially compared to larger datasets like WikiMatrix. This suggests that improving multilingual knowledge alignment may be better achieved through a mixed-language approach at the sample level rather than by incorporating simple translation data. 

\begin{figure}[t!]
            \centering
            \includegraphics[width=1\linewidth]{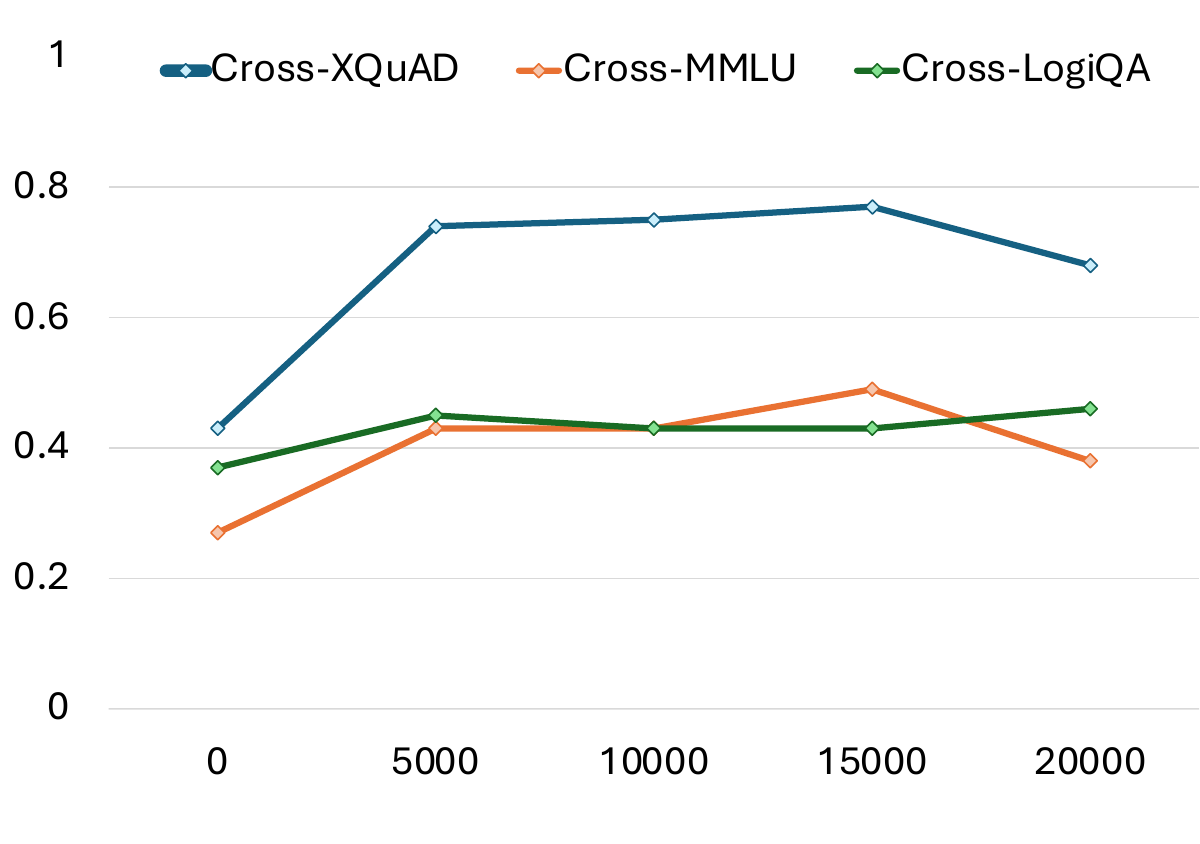}
            \caption{Comparison of AC3 score by adding different numbers of \texttt{CrossIn} data. Base model: \emph{Mistral-7B-v0.1}}
            \label{fig:quantity}
        \end{figure}

        \noindent\textbf{Essential Cross-Lingual Data Quantities.}         
        Figure~\ref{fig:quantity} shows the AC3 score of the LLMs with different quantity of cross-lingual alignment data. It can be shown that adding 5000 alignment data could already achieve a good result of cross-lingual consistency, there are not much improvement trend if we add more data. The observation that only a small amount of cross-lingual alignment data is required to achieve satisfactory consistency in LLMs can be attributed to its efficient learning mechanism. This characteristic allows the model to quickly assimilate and generalize from limited data, making it particularly adept at few-shot learning scenarios. Additionally, the model's pretraining on diverse linguistic corpora might have already equipped it with a foundational understanding of various languages, thereby reducing the need for extensive alignment data to bridge linguistic gaps. This efficient use of data not only demonstrates the model's robustness but also highlights its practicality in situations where data availability is constrained.

\section{Conclusion}

    In this paper, we presented a study on improving cross-lingual knowledge alignment of multilingual large language models, and contributed to both evaluation benchmarks and methodologies. We built a machine comprehension dataset that is a robust resource for extensive multilingual evaluation, emphasizing cross-lingual consistency in compensation with previous datasets. Our cross-lingual instruction tuning method \texttt{CrossIn} brought significant improvements in knowledge accuracy and consistency across languages, highlighting the potential of efficient tuning to create more robust multilingual large language models.

\section*{Limitations}

Our approach depends on the availability of high-quality translation and cross-lingual data, which may not be accessible for all languages. Addressing these data availability challenges is essential for further research on enhancing multilingual consistency in large language models.

In this study, we did not examine the impact of our cross-lingual data formulation on the pretraining stage of large language models. Pre-training is crucial as it significantly shapes the model's foundational knowledge and capabilities. Considering the larger scale of pretraining compared to fine-tuning, exploring whether our method could improve the efficiency and effectiveness of pretraining multilingual language models is a vital direction for future research. However, conducting such an ablation study on the pre-training stage is computationally demanding and may not be feasible with limited resources.

\bibliography{anthology,custom}

\appendix
\clearpage

\section{Appendix}
\label{sec:appendix}

\subsection{Prompt for Building Cross-XQuAD Data}

\begin{figure}[h!]
    \centering
    \includegraphics[width=1\linewidth]{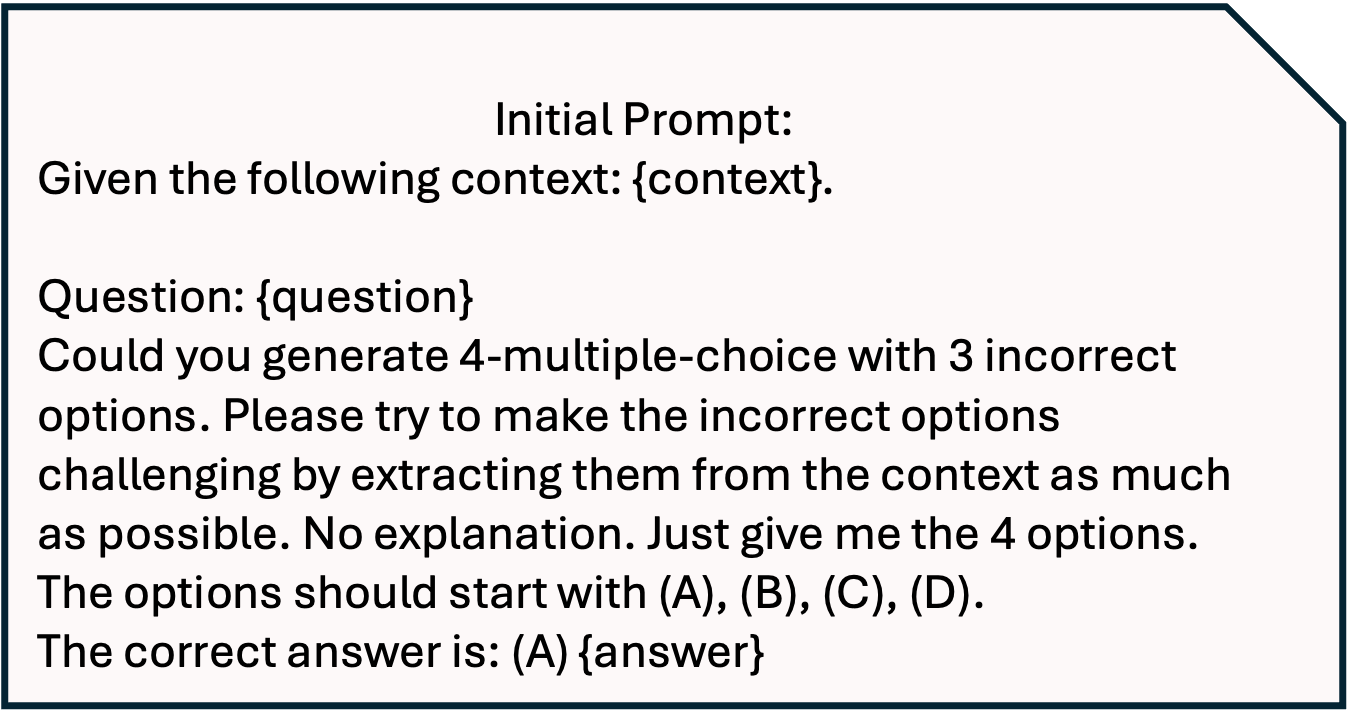}
    \caption{Prompt For Generating English Choice}
    \label{fig:create-prompt}
\end{figure}

\begin{figure}[h!]
    \centering
    \includegraphics[width=1\linewidth]{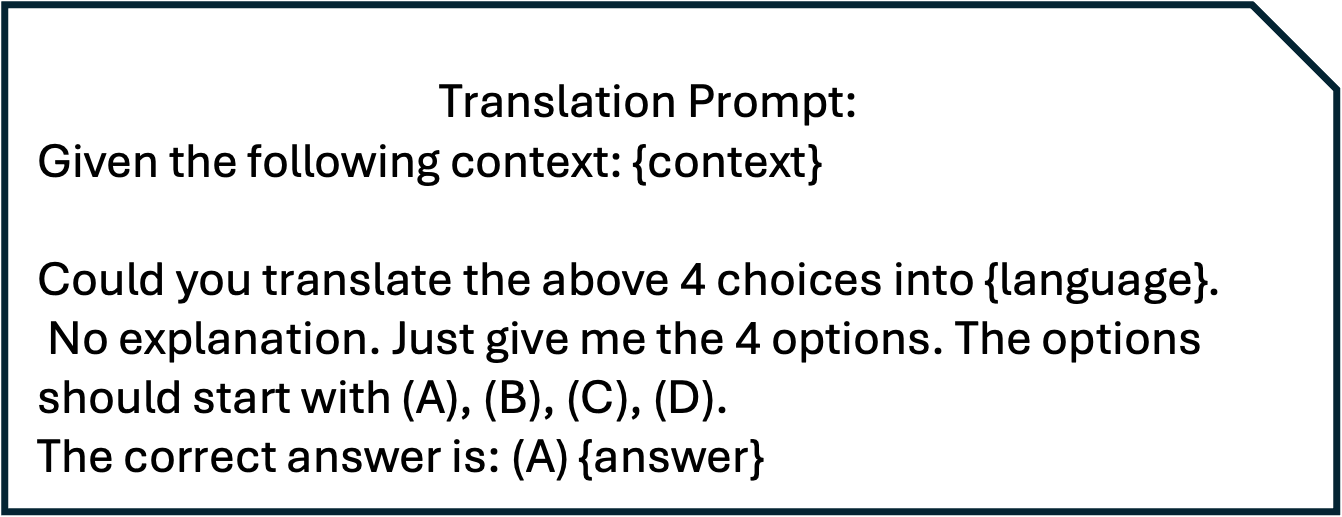}
    \caption{Prompt to Translate English Choice}
    \label{fig:translate-prompt}
\end{figure}

\subsection{Fine-tuning Parameters}

\begin{table}[ht]
\centering
\label{tab:hyperparameters}
\begin{tabular}{@{}lp{5cm}@{}}
\toprule
\textbf{Hyperparameter}     & \textbf{Value}     \\ \midrule
learning\_rate      & 1e-4      \\ \hline
batch\_size         & 16        \\ \hline
epochs              & 1         \\ \hline
lora\_rank          & 64        \\ \hline
lora\_alpha         & 128       \\ \hline
lora\_trainable     & p\_proj, k\_proj, v\_proj, o\_proj, gate\_proj, down\_proj, up\_proj \\ \hline
modules\_to\_save   & embed\_tokens, lm\_head \\ \hline
lora\_dropout       & 0.05      \\ \hline
warmup\_ratio       & 0.03      \\ \hline
weight\_decay       & 0         \\ \hline
optimizer           & Adam      \\ \hline
bf16                & True      \\ \hline
\bottomrule
\end{tabular}
\caption{Fine-tuning Hyperparameters}
\end{table}

\end{document}